\documentclass[10pt, sigconf]{acmart}

\usepackage{multirow}
\usepackage[linesnumbered,ruled]{algorithm2e}
\usepackage{algpseudocode}
\usepackage{citeall}
\usepackage{amssymb}
\usepackage{amsmath}
\def\BibTeX{{\rm B\kern-.05em{\sc i\kern-.025em b}\kern-.08emT\kern-.1667em\lower.7ex\hbox{E}\kern-.125emX}}
    
\copyrightyear{2019}
\acmYear{2019}
\setcopyright{acmlicensed}
\acmConference[CIKM'19]{Proceedings of CIKM'19}{November 03--07, 2019}{Beijing, China}
\acmBooktitle{Proceedindgs of CIKM'19, November 03--07, 2019, Beijing, China }

\begin{document}

%
\title{A Machine Learning Framework for Data Ingestion in Document Images}

%
\author{Han Fu$\dagger$, Yunyu Bai$\dagger$, Zhuo Li$\ddagger$, Jun Shen$\ddagger$, 	Jianling Sun$\dagger$
}
\affiliation{
	\institution{$\dagger$Zhejiang University, Hangzhou, China}
	\institution{$\ddagger$State Street Corporation, Hangzhou, China}
}

%
\renewcommand{\shortauthors}{ }

%
\begin{abstract}
Paper documents are widely used as an irreplaceable channel of information in many fields, especially in financial industry, fostering a great amount of demand for systems which can convert document images into structured data representations. 
In this paper, we present a machine learning framework for data ingestion in document images, which processes the images uploaded by users and return fine-grained data in JSON format. Details of model architectures, design strategies, distinctions with existing solutions and lessons learned during development are elaborated. We conduct abundant experiments on both synthetic and real-world data in State Street. The experimental results indicate the effectiveness and efficiency of our methods.

\end{abstract}

%
%


%

%

%
\maketitle

\section{Introduction}
Information ingestion from structured documents has drawn considerable attention from both research and application area of data mining and information retrieval. The substantial variability of layouts and data contents makes it difficult to precisely convert the content into structured representations. Further, it is more challenging when the document contents are not encoded in editable data formats (e.g. PDF) but are captured as images. Document images are common in business world, such as fax, bills and receipts, etc. Paper documents are widely used as a essential information medium, especially in government departments and financial industry for security reasons. Naturally, there comes a great demand to automatically convert contents of document images into structured representation formats. With the rapid development of machine learning techniques, many methods have been proposed for document layout analysis \cite{sunder2019one,augusto2017fast,raoui2017deep}. However, to our knowledge, there is still not a technique to meet the demand of industrial application due to following challenges:

\begin{figure*}[h]
	\centering
	\includegraphics[width=12cm]{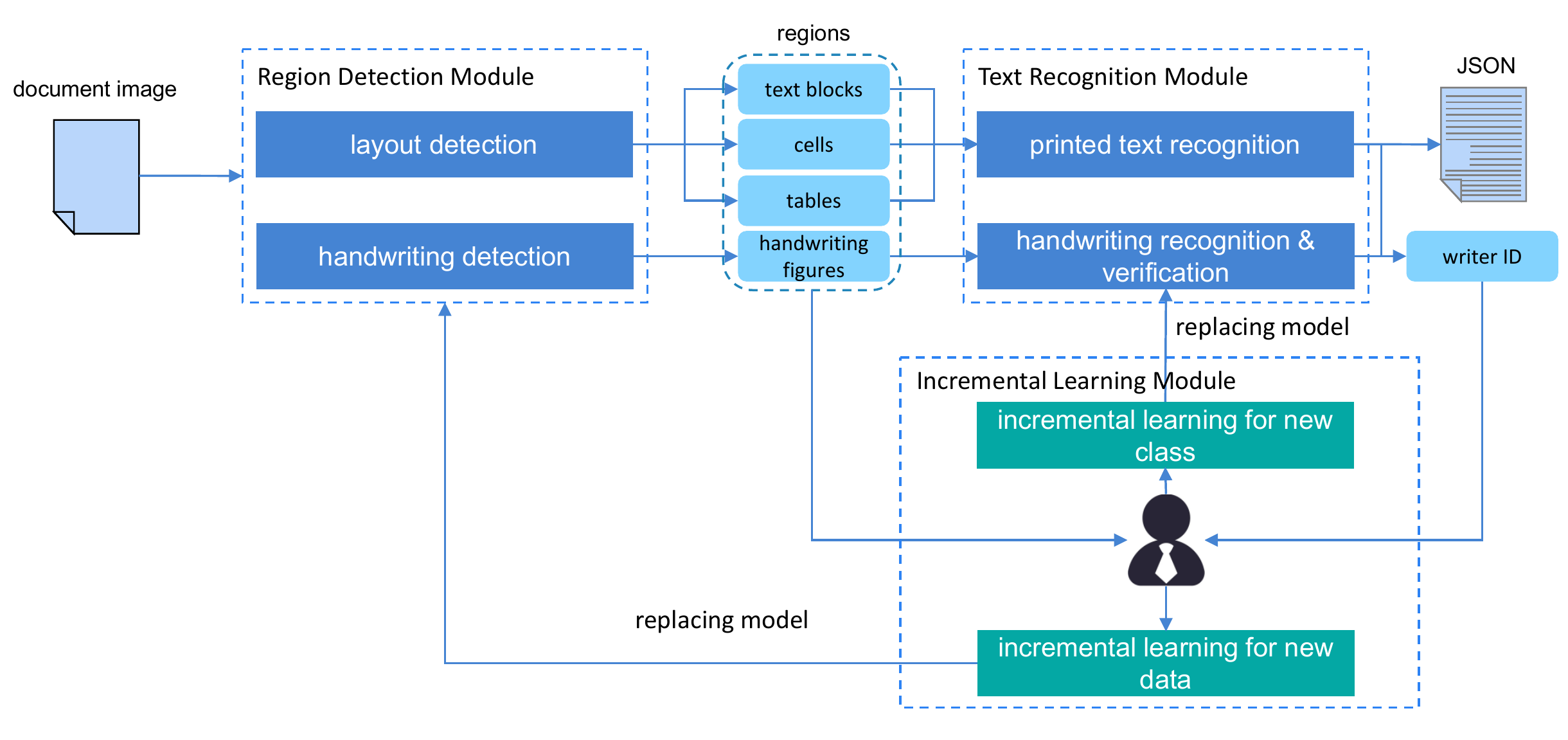}
	\caption{Workflow of data ingestion from document images. The approach takes a document image as input and outputs structured JSON containing fine-grained textual data. }
	\label{figure1}
\end{figure*}

\begin{itemize}
\item[(1)]A document image consists of multiple regions with different semantic labels, such as tables, figures, signatures, text blocks, etc. Optical Character Recognition (OCR) methods are not capable to capture the differences between these regions.
\item[(2)]The precision of current document analyzers decreases sharply when solving tabular data without form lines.
\item[(3)]The accuracy of handwriting verification and recognition can hardly meet the standard for industrial use.
\end{itemize}
Consequently, the task of information extraction from document images is often manually done.

Motivated by such challenges, we propose a generic framework for document image understanding systems to reduce human intervention. Specifically, the framework involves following sub-modules as shown in Figure \ref{figure1}:

\begin{itemize}
	\item \textbf{The region detection module} segments a document image page into several document regions and predict the corresponding semantic labels. In this paper, we only focus on handwritings, tables, tabular cells and text blocks.
	\item \textbf{The text Rrcognition module} takes the location information of each region as input and outputs text sequences. To reduce coupling, we train independent models to recognize handwritings and printed texts.
	\item \textbf{The incremental learning module}: this module is responsible for fine-tuning the trained models to adapt to new classes or new data.
\end{itemize}

To improve the efficiency and effectiveness of our system, we propose multiple novel models and training algorithms for each task. Details of design ideas, system architectures and model settings are provided in this paper. The entire system is currently deployed in the development environment of State Street Corporation but not yet in production.
 
The rest of this paper is structured as follows: Section 2 introduces state-of-the-art document understanding solutions. In section 3, we present the overall design of the data ingestion process and architecture of each components. Section 4 presents the overall system design and pipeline of document understanding process. In section 5, we present the experimental results and discuss the lessons learned during model training and system developing. Last but not the least, we conclude the paper and discuss relative open questions and possible next steps for this solution in section 6.

\section{Related Works}
The problem of data extraction from documents images has drawn attention over decades \cite{nagy2000twenty}. Generally, it involves two types of tasks: text recognition and document parsing. With the application of deep neural networks, OCR engines such as Tesseract \cite{smith2007overview} have achieved good performance in dealing documents with simple textual layouts. \cite{jaderberg2014synthetic} firstly employ convolutional neural networks to extract text features and train end-to-end learnable recognition models. Following this work, several advanced approaches were proposed. Broadly speaking, these approaches can be categorized into two classes: separated text detection and recognition processes \cite{shi2017end, tian2016detecting, he2017deep, lyu2018multi, borisyuk2018rosetta}, or joint text detection and recognition processes such as \cite{li2017towards, buvsta2017deep, liu2018fots}. However, document images have natural structured regions to represent various data, such as tables, figures and text blocks. One single OCR step is not enough for such analysis problem. Consequently, there are some works concentrating on extracting document data directly into structured formats \cite{cesarini1998informys, chanod2005legacy, peanho2012semantic, li2016precomputed}. However, such works usually need accurate textual information and expertise knowledge from users, which limits their adaptability in industry applications. Moreover, \citeauthor{raoui2017deep} and \citeauthor{augusto2017fast} recently propose to leverage deep convolutional networks to recognize different regions of document images. However, these approaches can hardly handle document images with complex data representations, such as tables without form lines. Different from the related methods, we propose to formulate the layout recognition problem as a object detection task. With various scale of view, out systems is capable to locate each table cell, and the tabular contents will be simply converted to structured representations with a table construction algorithm. Object detection is a hot research topic in computer vision area and state-of-art approaches \cite{ren2015faster, dai2016r, he2017mask} have been utilized for words detection \cite{borisyuk2018rosetta, tian2016detecting, sunder2019one}. In this work, we use Light-Head RCNN \cite{li2017light} for both precision and efficiency consideration.

Besides the high performance, the reason why we choose deep neural network-based methods in our system is that, the models are expected to be fine-tuned to quickly adapt to the new coming data or new class, which is natural for a online system. There exists several researches working on incrementally improving trained models with online data such as \cite{shmelkov2017incremental, su2016line}. Inspired by such works, we design an interactive learning strategy to train online models with author feedbacks and avoid the catastrophic forgetting problem.

\begin{figure*}[h]
	\centering
	\includegraphics[width=15cm]{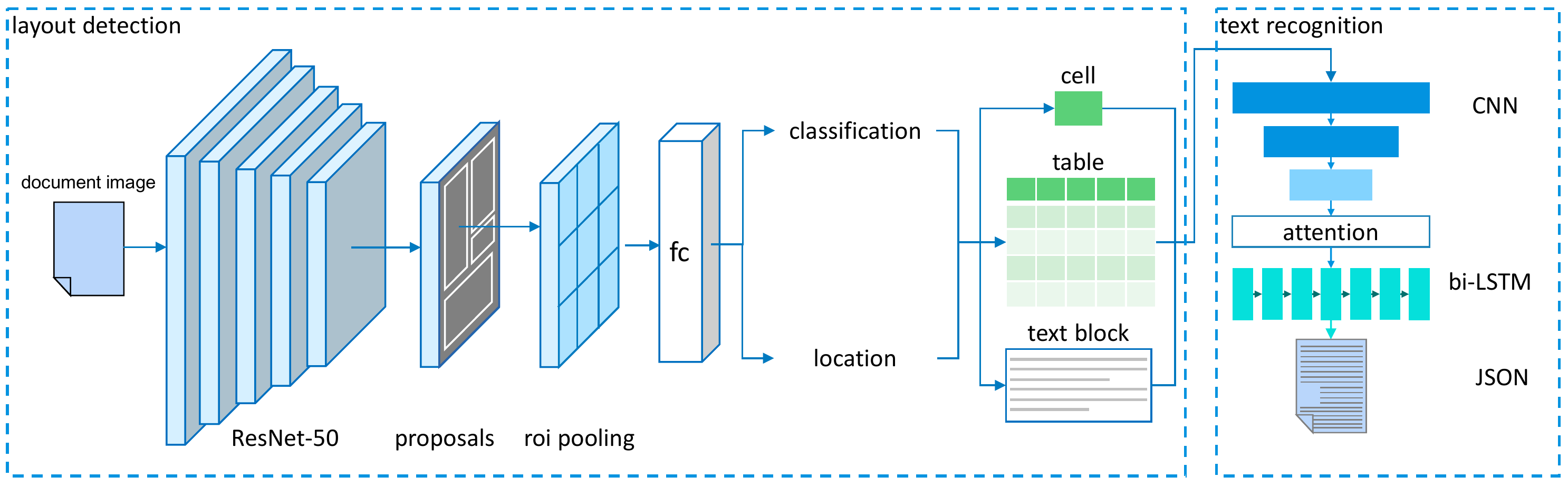}
	\caption{A diagram of data ingestion process for printed texts. The architecture of region detection model is on the left, and the right part comes the recognition model for printed texts.}
	\label{figure2}
\end{figure*}

\begin{figure}[h]
	\centering
	\includegraphics[width=5cm]{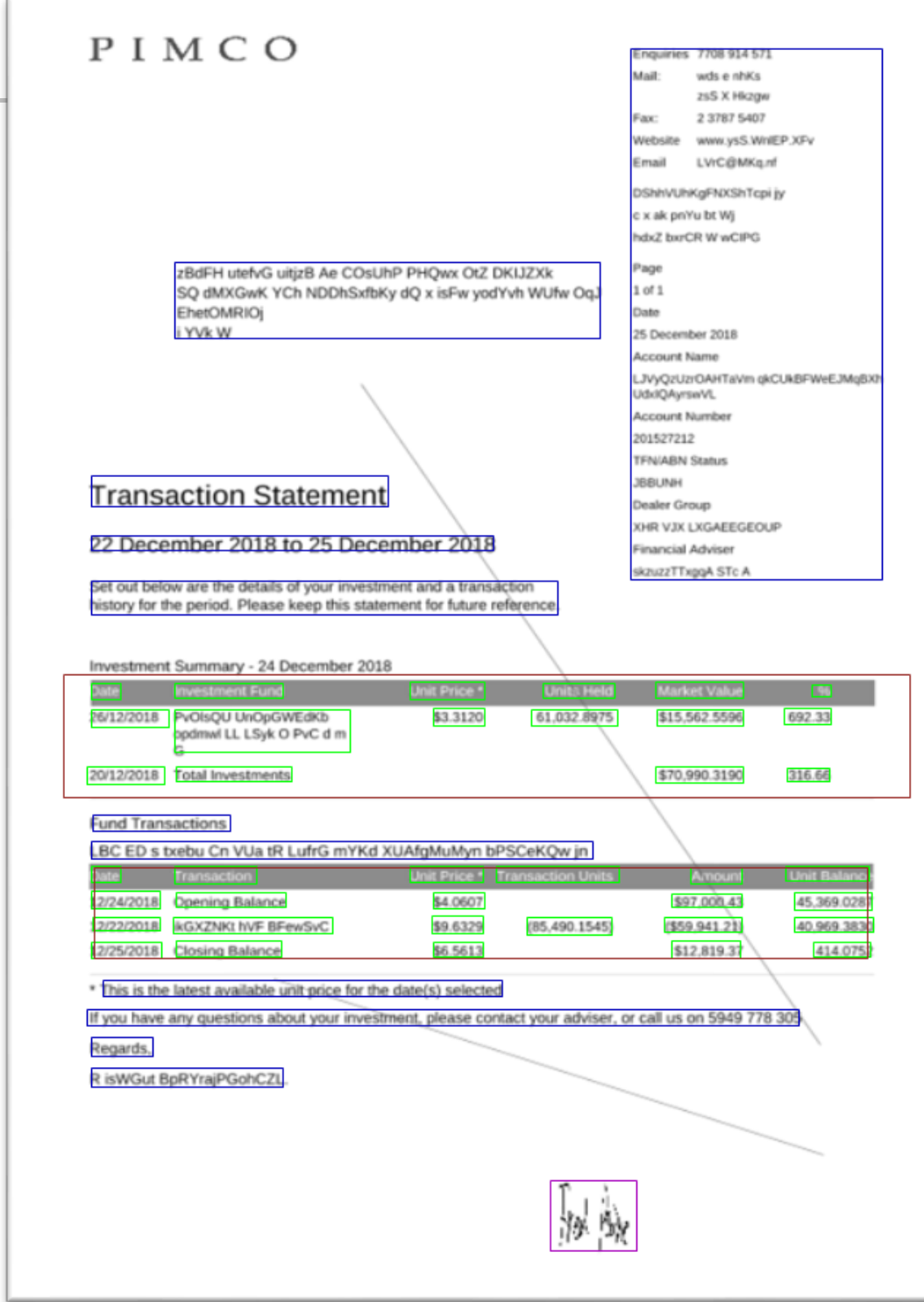}
	\caption{Regions detected on a sampled document image, where the tables, tabular cells, text blocks and handwritings are labeled with red, green, blue and purple boxes.}
	\label{figure5}
\end{figure}

\section{Data Ingestion from Document Images}

In this section, we introduce the proposed data ingestion approaches and point out how our methods differ from the existing solutions. 

As illustrated by Figure \ref{figure1}, the system comprises of three major function components: 1) a region detection module, 2) a text recognition module and 3) a incremental learning module. 

Given a document image, the first step is to detect rectangle regions of data, and predict the corresponding layout labels including handwriting blocks, tables, tabular cells and text blocks. In the second step, we apply text recognition, where, the detected regions are fed to a attention-based convolutional recurrent (ACRNN) model to recognize the texts. This workflow enjoys several advantages. First, once the regions are detected, text recognition can be performed in parallel over all regions. Second, the detection module and recognition module can be updated independently. Furthermore, since the models are decoupled, they can be shared by other systems as individual interfaces.

Moreover, different from existing solutions, the machine learning models in this system are not statically pre-trained before deployment, but can be dynamically updated according to new data or added classes. To this end, we propose a knowledge distillation-based incremental learning module.

\begin{figure*}[h]
	\centering
	\includegraphics[width=15.3cm]{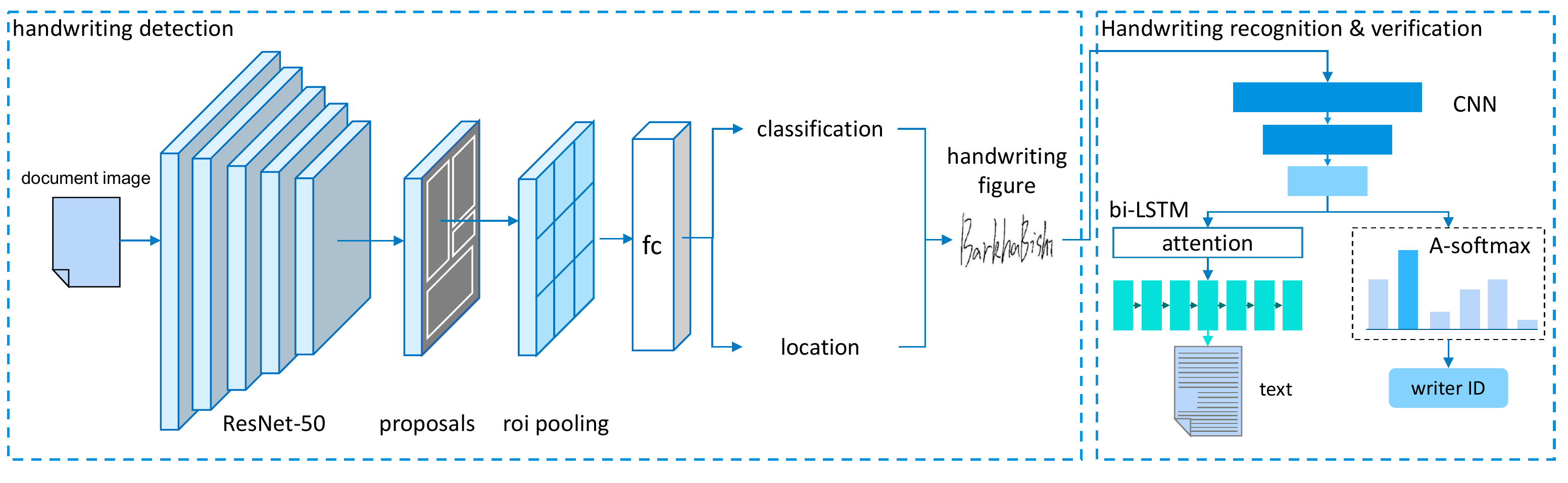}
	\caption{The architecture of handwriting detection model (left) and recognition model (right). The handwriting recognition model is trained to jointly learn sequence recognition and handwriting verification simultaneously.}
	\label{figure3}
\end{figure*}

The whole process of document understanding is conducted semi-automatically as it is too risky to allow full straight through processing in document ingestion when the image quality is impacted by many factors. A UI is provided for operational staff to review and edit results. At every step where manual check is involved, any correction made will be fed into relative incremental training process to continuously refine the model.

\subsection{Region Detection Module}
We train two independent models to detect handwritings and other regions (tables, cells, and text blocks) respectively for the purpose of task decoupling. Another important reason is that it is hard to train a single model to capture heterogeneous features of handwriting and printed texts simultaneously. For ease in writing, we refer the process of locating tables and text blocks as \emph{layout detection}. An example of regions we are interested in is shown in Figure \ref{figure5}.

\subsubsection*{\rm \textbf{Handwriting Detection}}

For handwriting detection, we build a object detection model based on Light Head R-CNN \cite{li2017light}. The model is trained on a two class problem, namely, "handwriting" and "background". The detection process can be summarized as following steps as shown in Figure \ref{figure3}, 1) processing the entire image with a ResNet-based basic feature extractor \cite{he2016deep} and a large separable CNN \cite{szegedy2016rethinking} sequentially to extract the feature representation, 2) generating several regionals of interest (RoI) via a region proposal network (RPN) by feeding the feature map as input, and 3) leveraging a R-CNN subnet to predict class label and location of each region candidate. Our detection model uses typical architecture of Light Head R-CNN but replaces the ResNet-101 CNN extractor with ResNet-50 for efficiency. The entire model is trained end-to-end using stochastic gradient methods.

\subsubsection*{\rm \textbf{Layout Detection}}

To construct a structured data representation, we need to find the bounding boxes of each minimum area with a independent semantic label, such as a text block or a tabular cell. However, since texts in tables and other snippets often share fonts and sizes, it is extremely challenging to capture all such regions by training a single model. Concretely, all existing solutions segment a document image into tables and text blocks, but fail to provide more find-grained data representations. In this work, we address this challenge by formulating the layout detection task as a four class problem, including text blocks, tabular cells, tables and the background. The introduction of class \emph{table} can improve the detection accuracy of cells by explicitly providing
the relative spatial relationship between tables and text blocks. The model architecture is similar to that of the handwriting detection model by adopting Light Head R-CNN on a four class problem as shown in the left part of Figure \ref{figure2}. Moreover, since the layout detection model should be able to recognize both coarse-grained (tables and text blocks) and fine-grained regions (tabular cells), we modify the original RPN in Light Head R-CNN to generate anchors of wider range of scales. To be specific, the tweaked RPN generates 56 anchor boxes for each RoI with eight scaling ratios ($\{0.25, 0.5, 1, 2, 4, 8, 16, 32\}$) and seven aspect ratios ($\{0.25, 0.33, 0.5, 1, 2, 3, 4\}$). Once the final regions are determined, we apply a cell combination algorithm to construct table-structured data representations. This algorithm constructs rows and columns according to y-coordinates and x-coordinates respectively with the similar manner. The algorithm for data rows is detailed in Algorithm \ref{algorithm3}. Once all regions are extracted, the location information is constructed as a JSON file, ordered by y-coordinate of the top edge.

\begin{algorithm}[htb]
	\caption{Table Construction for Data Rows}
	\label{algorithm3}
	\raggedright
	\KwIn{$\mathcal{C}=\{c_1,\cdots,c_N\}$, where
		$\mathcal{C}$ is the list of tabular cells sorted by the y-coordinates of the top edges; the height threshold $H$.
	}
	\KwOut{$\mathcal{R}=\{r_1,\cdots,r_N\}$, where $r_i$ denotes the set of rows that $c_i$ occupies.}
	$n=0$;\\
	$y\_last=-\infty$;\\
	\For{$i=1$; $i<=N$; $i++$}{
		Get the y-coordinates of top and bottom edge of $c_i$: $ymin_i$, $ymax_i$;\\
		\uIf{$ymin_i-y\_last>H$}{
			$n++$;\\
			$y\_last=ymax_i$;\\
			Add $n$ to $r_i$;\\
			\ForEach{\rm{$c_j$ in Row $n-1$}}{
				\If{$ymax_j$ > $ymin_i$}{
					Add $n$ to $r_j$;
				}
			}
		}
		\Else{
			Add $n$ to $r_i$;
		}
	}
\end{algorithm}

\subsubsection*{\rm \textbf{Handling Overlapped Regions}}

Since several RoI can be overlapped, non-maximum suppression (NMS) is employed by typical object detection models to reduce redundant proposals. For handwriting blocks, we simply set the intersection-over-union (IoU) threshold of NMS as 0.3 to reduce overlapped region candidates. However, the situation is complicated when applying layout detection. We observe that one single tabular cell or text block is usually detected as several overlapped regions due to the existence of spaces and empty row, which severely harms the data integrity. And simply setting the IoU threshold to a small value will lead to missing valid regions. To address this problem, we propose a region combination approach (RC) by extending the standard NMS. Specifically, the first step is to apply standard NMS with a large IoU threshold (0.7 in this work) to reserve as many candidate regions as possible. In the second step, all overlapped rectangles are combined as one. The process is detailed in Algorithm \ref{algorithm2}. 

\begin{algorithm}[htb]
	\caption{Region Combination Algorithm}
	\label{algorithm2}
	\raggedright
	\KwIn{The set of initial candidate regions $\mathcal{H}$ of the same class.}
	
	\KwOut{The set of final regions $\mathcal{L}$ initialized as $\varnothing$.}
	Reduce overlapped region proposals by applying NMS with a large IoU threshold (0.7 in this work). The intermediate set of all reserved regions is denoted by $\mathcal{M}$;\\
	\While{$\mathcal{M}\neq\varnothing$}{
		Find the region $r$ from $\mathcal{M}$ with the highest confidence score;\\
		Remove $r$ from $\mathcal{M}$;\\
		\ForEach{\rm{region candidate} $c$ \rm{in} $M$}{
			\If{$IoU(r, c)>0$}{
				Combine $r$ and $c$, and the combined region is denoted by $s$;\\
				Add $s$ to $\mathcal{N}$;\\
				Remove $c$ from $\mathcal{M}$;
			}
		}
	}
\end{algorithm}

\subsubsection*{\rm \textbf{Training Strategy}}
\begin{figure}[h]
	\centering
	\includegraphics[width=\linewidth]{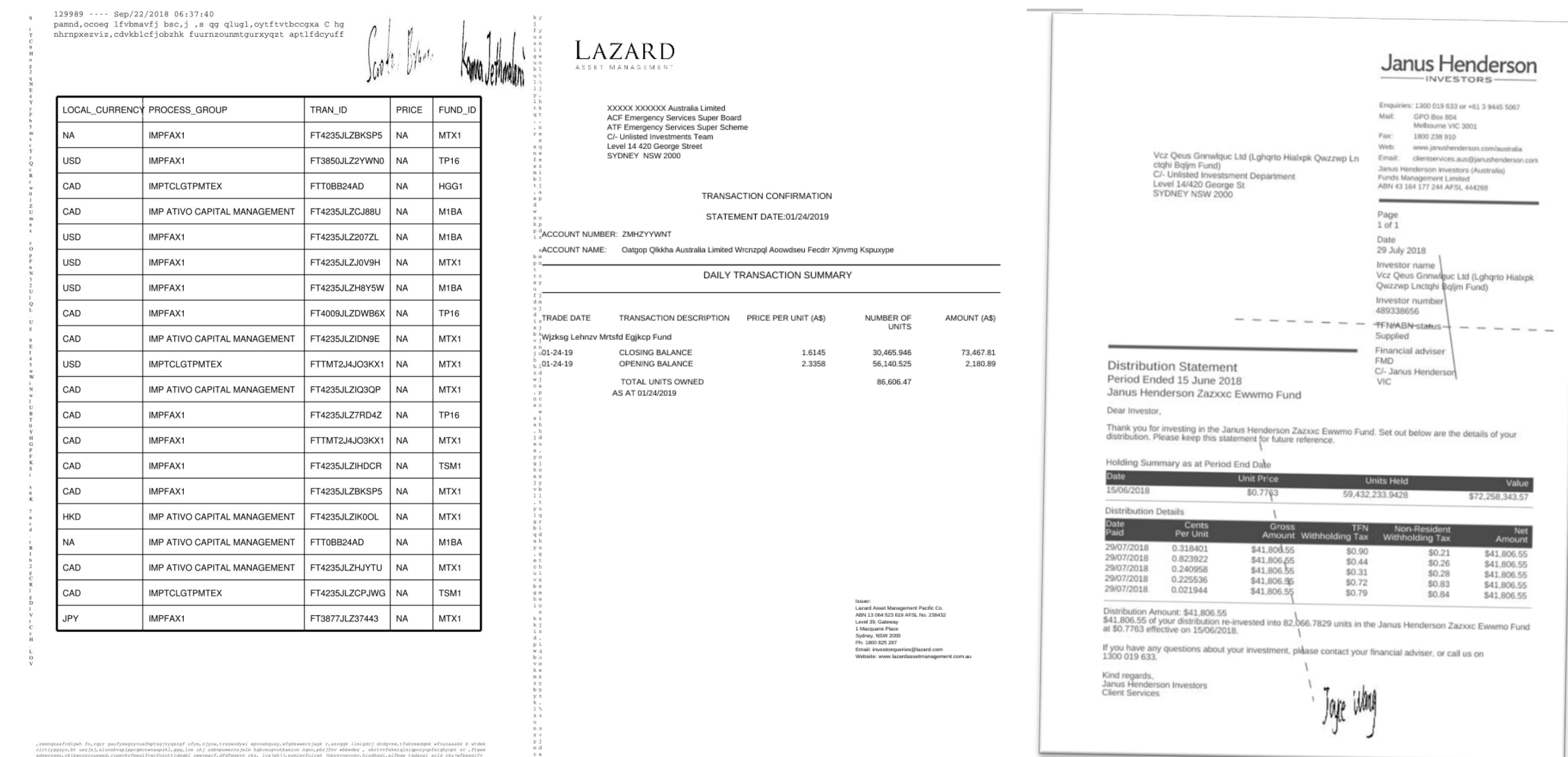}
	\caption{Comparison between real scene documents (middle and right) and a synthesized one. }
	\label{figure4}
\end{figure}

The handwriting detection model can be trained end-to-end directly. In the contrary, the layout detection model is hard to train on document images with complicated styles. To address this challenge, we propose a warm-up training strategy to pre-train the model on synthesized data, inspired by \cite{bengio2009curriculum}. For model training's efficiency and effectiveness concern, we artificially synthesize a large amount of mock data, which has similar data structures (e.g., tables and text blocks) with the real document images but the visual styles (e.g., fonts and shading) are much more simplex. Once the pre-trained model is converged, we set it as the initialization weights and further train the detection model on document images which are collected in the real scenario with sensitive information hidden. Such document images are named as \emph{true data} for convenience in the rest of this paper. This easy-to-hard training strategy makes model converging process faster. \ref{figure4} shows both an example synthesized document image and samples from the true data set.

\subsection{Text Recognition Module}
Once the regions are detected, the relative location information is then passed to the text recognition module which extracts characters sequentially from the regions. Similar to the decoupling strategy implemented in the region detection module, we also train two independent models to recognize printed texts and handwritings respectively. 

\subsubsection*{\rm \textbf{Printed Text Recognition}}
Typically, a recognition system for printed texts involves two components, a line detector and a character extractor. The SOTA document image analyzers \cite{sunder2019one,le2019deep} use deep CNN-based detection models such as Fully Convolutional Network (FCN) \cite{zhang2016multi} and Connectionist Text Proposal Network (CTPN) \cite{graves2006connectionist} to detect text lines. Different such solutions, we utilize the Tesseract line segmentation algorithm \cite{smith2007overview,smith1995simple} based on connected component analysis \cite{rosenfeld1976digital}. The reasons can be summarized as two points. First, we have already predicted the rectangles which precisely cover each text block or tabular cell. With this step, we can assume that all characters in a region share the same font and size. Therefore, the text lines can be segmented precisely according to the blank areas. The second point is that CNN-based models undoubtedly require more computation resource and inference time. 

After text line segmentation would follow a sequence recognition model. Since the length of text in a line is not fixed, the recognition model should be capable to handle sequences with arbitrary lengths. Existing solutions \cite{borisyuk2018rosetta,li2017towards} employ Connectionist temporal classification \cite{shi2016end} output layer or lstm encoder-decoder architecture to address this problem. Different from these works, we only use a convolutional network to extract sequential visual features and an attention-based lstm decoder character extraction as shown in Figure \ref{figure2}. We refer this model as attention-based convolutional recurrent neural network (ACRNN) in the rest of this paper. Considering that document images are usually clear and easy to decipher compared with scene texts, we choose a much smaller architecture for efficiency. Specifically, we use Google BN-Inception \cite{szegedy2016rethinking} as the basic feature extractor and the dimension of lstm hidden state is set as 256.

\subsubsection*{\rm \textbf{Handwriting Recognition}}
Handwriting recognition is a challenging research topic which has been studied for decades. There are two major distinctions between printed texts and handwritings. First, the features of handwritten texts are influenced by the writing styles of different individuals. Second, since the input images of a text recognition model should be resized to the same width during training, the length of output sequence therefore depends on various writing habits. Motivated by these factors, we propose a novel handwriting recognition model, which jointly learns to extract characters from handwriting regions and verify the handwritings simultaneously as shown in Figure \ref{figure3}. In a nutshell, this method consists of two sub-models sharing the same CNN feature extractor. We incorporate a image classification model into the original sequence recognition model, which is responsible for identifying the owner of handwritings. The output layer of the classification model is Angular Softmax (A-softmax) \cite{liu2017sphereface}, which makes the decision boundary more stringent and separated. In this way each handwriting can be more likely to be categorized into a certain class rather close to the boundary between two classes. Formally, the training object of the whole model is:
\begin{equation}
\label{eqn_multi}
\min \lambda\mathcal{L}_{c} + \frac{1}{N}\mathcal{L}_{r},
\end{equation}
where $\mathcal{L}_{c}$ and $\mathcal{L}_{r}$ are the loss functions of handwriting and sequence recognition respectively, $N$ is the length of the target text, and $\lambda$ controls the ratio between two training losses. In this work, we simply set $lambda$ to 1. This model enjoys several benefits, including the ability to distinguish the handwriting features belonging to different individuals, and improve the robustness with multi-task learning. In the rest of this paper, we refer this multi-task model as ACRNN-MT.

Moreover, the incorporation of handwriting verification brings additional application value. In practice, many financial document images, such as fax, receipt, invoice, contain signatures and handwritten modifications, where the handwritings must be verified to guarantee the data security.

\subsubsection*{\rm \textbf{Image Preprocessing}}
Image preprocessing plays a critical role in text recognition. Following current approaches to text recognition \cite{jaderberg2014synthetic}, we resize the input images to 32x128 pixels. This process is necessary to train the models in parallel with batches of images. Straight after reshaping the inputs, we apply several preprocessing methods to improve image quality, including grayscaling, brightness enhancing, background cleaning, noise reducing, sharpening, trimming background, and border padding. All these functions are provided by open source ImageMagick \cite{imagemagick2014imagemagick}.

\subsection{Incremental Learning Module}
One of the crucial distinctions between SOTA systems and ours is that the machine learning models are not \emph{static}. That means, we do not only apply pretrained models for different tasks, but also fine-tune the models to adapt for new coming data or additional class labels. In this paper, we focus on two cases:
\begin{itemize}
	\item[(1)] Region detection models generate inaccurate bounding boxes on new coming images and users have provided corresponding corrections.
	\item[(2)] For the handwriting verification model, new classes should be added when new writers are authorized. 
\end{itemize}

The above tasks can be formulated as transfer learning or domain adaptation problems. Ideally, the models are expected to be fine-tuned to adapt to new data or new class. However, neural networks may suffer from \emph{catastrophic forgetting} in the absence of the original training data. The internal representations of neural networks can be severely affected by domain vicissitude of the new coming data. Motivated by such challenges, we propose a knowledge distillation-based incremental learning method inspired by \cite{shmelkov2017incremental}. 

Specifically, given a trained model $M$ with fixed parameters, the first step is to make a copy $M^{*}$ from $M$. We modify the output layer of new copied model only if new classes of handwritings are added, and keep the model unchanged in other cases. At each timestep during training, we first sample a batch of data $T_{o}$ from $D$. Then we divide both $M$ and $M^{*}$ into several groups and compute the distillation loss on $T_{f}$ as:

\begin{equation}
\label{eqn_dist}
\mathcal{L}_{dist} = \sum_{l=1}^{L}||\phi_{M}^{l}(T_{o})-\phi_{M^{*}}^{l}(T_{o}) ||_{2}^{2},
\end{equation}
where $||\cdot||$ is L2 norm, $\phi_{M}^{l}$ and $\phi_{M^{*}}^{l}$ denotes the output layer of the $l$-th group of $M$ and $M^{*}$ respectively. For object detection models, the groups are divided according to the function of different components, including feature extractor, region proposal network and R-CNN subnet. The feature extractor CNN is further separated into several blocks with respect to the pooling processing (BN-Inception) or residual blocks (ResNet). To adapt the model for the new task, we train $M^{*}$ on $T_{f}$ by minimizing the standard classification or detection loss $\mathcal{L}_{task}$. The final loss is computed as:

\begin{equation}
\label{eqn_dist_loss}
\mathcal{L}=\mathcal{L}_{task} + \alpha\cdot\mathcal{L}_{dist}
\end{equation}
where $\alpha$ is a hyper-parameter to control the ratio between the distillation loss and original loss, which is simply set to 1 in this work. The algorithm is detailed in Algorithm \ref{algorithm1}.

\begin{algorithm}[htb]
	\caption{Incremental Learning to adapt to new data or new class.}
	\label{algorithm1}
	\raggedright
	\KwIn{Trained Model $M$; Author feedback set $D_{f}$; Original training dataset $D$; Batch size $B$; New data rate $p$; Weight of distillation loss $\alpha$, Training algorithm $A$}
	
	\KwOut{Trained New Model $M^*(\theta)$}
	\While{not converge}{
		Random sample a training batch $T$, including $B\times p$ samples from $D_f$ as $T_{f}$ and $B\times (1 - p)$ samples from $D$ as $T_{o}$ respectively \\
		Compute distillation loss $\mathcal{L}_{dist}$ on $T_{o}$ as Eqn. \ref{eqn_dist}\\
		Compute task loss $\mathcal{L}_{task}$ on $T_{f}$ \\
		Update model paramter $\theta$ with $A$ by minimizing training loss $\mathcal{L}=\mathcal{L}_{task} + \alpha\cdot\mathcal{L}_{dist}$
	}
\end{algorithm}

\begin{figure}[h]
	\centering
	\includegraphics[width=8.5cm]{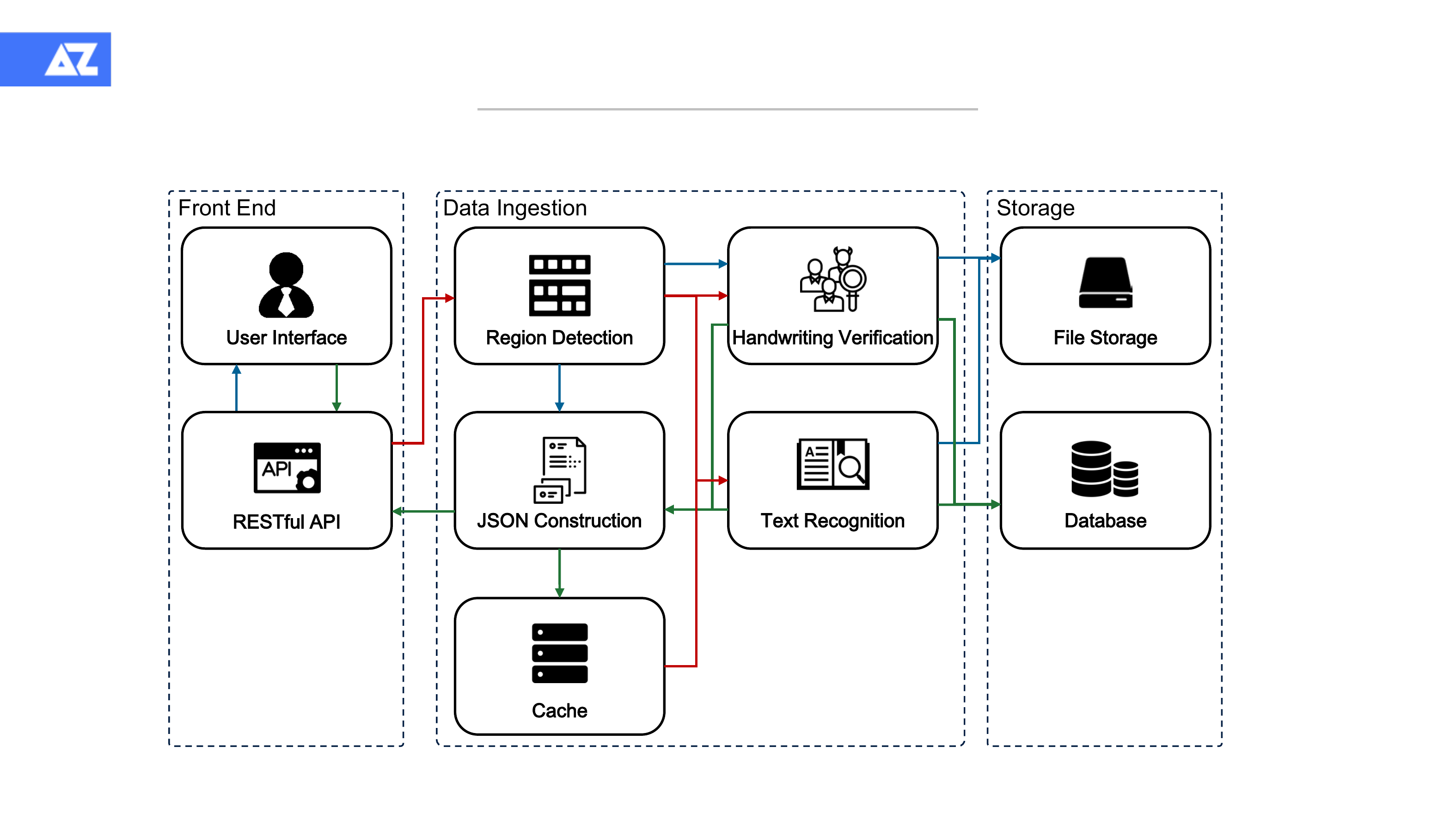}
	\caption{Architecture of the document understanding system.}
	\label{figure6}
\end{figure}

\section{System Architecture}
In this section, we introduce the architecture of the document image understanding system, which has been deployed in the development environment but not yet in production. As shown in Figure \ref{figure6}, the system consists of three components, a front end server providing a user interface, a back end for computation, and a storage layer for data persistence. The entire process of document image understanding is as follows:

\begin{itemize}
	\item[(1)] The user uploads a image through a user interface, and then the image is downloaded to the back-end server.
	\item[(2)] The image is pre-processed and passed to the region detection module. The location information is constructed as JSON format and returned to UI through a RESTful API. The user is required to check the returned results and correct the mistakes. The JSON data is also written to a in-memory cache, and the raw image is stored in the file storage. In this work, we use Redis\footnote{https://redis.io/} as the in-memory storage.
	\item[(3)] The text recognition module reads JSON data from the cache and fill in the recognized texts and verification information. The parsed data is finally stored in the database for data persistence.
	\item[(4)] All correction information provided by the users is staged in the cache until the incremental learning is executed regularly.
\end{itemize}

To reduce the time spending in communication, we use coroutines for concurrence in the back-end, which is capable to serve at most 1024 manual checkers synchronously with little memory overhead. The detailed time cost of different machine learning models for a single image is listed in Table \ref{time}. It should be noted that text recognition is executed concurrently among all regions. From the table, we can observe that the average time to process one single document image is less than 5 seconds, which is much faster than a manual worker does and applicable for financial industry. The step taking most time is the text recognition process and the inference time depends on the text length.

\begin{table}
	\caption{Averaged time overhead of the machine learning models.}
	\label{time}
	\begin{tabular}{ccc}
		\toprule
		\textbf{Function Module} & Model & 
		Inference Time \\
		\midrule
		Handwriting Detection & Light-head R-CNN & 0.32 secs \\
		Layout Detection & Light-head R-CNN & 0.79 secs \\
		Handwriting Verification & A-softmax & 0.19 secs \\
		Handwriting Recognition & ACRNN & 1.15 secs \\
		Printed Text Recognition & ACRNN & 2.47 secs \\
		\bottomrule
	\end{tabular}
\end{table}

\section{Experiments}
Here, we present the experimental results and describe the lessons learned from model training and system development.

\subsection{Data Collection}
The dataset is constructed by collecting fax images from the real business scenario in State Street Corporation. We remove the sensitive information such as personal details, signatures and fund names. Typical examples of the images are shown in Figure \ref{figure4}. After preprocessing, we obtain 84,000 fax images for training, 18,000 for validation and 18,000 for test respectively. For convenience, we refer such a collection as the \emph{true data}.

Since the collected fax images do not contain handwritten signatures after preprocessing, we manually collected handwriting figures provided by the staff in State Street. To be specific, we manually collected 86,481 signatures written by 486 individuals. In order to get a robust model to precisely distinguish handwriting features from different individuals, we ask the writers not to sign their own names but to pick five to ten names from a given name list which includes 510 English names, 260 Chinese PinYin names and 264 Indian names, with lengths vary from 5 to 32 characters. All names are signed on a blank form with A4 size. Papers with signatures are sent to a scanner and the scanned signatures are split according to form lines. The average height and width of the signature images are about 140 and 400 pixels respectively. The scanned signatures are processed using background elimination, randomly zooming and twisting, and finally pasted to the fax images to replace the original ones. We split the signature dataset as 72833/5000/8648 for training/validation/test of the handwriting recognition task.

Furthermore, as discussed above,  we use an easy-to-hard training strategy to warm up the training process with a large synthetic dataset.  The images in the synthesized dataset are much simpler in terms of layouts, but vary more in density of table/text cells, fonts and sizes. Since the features in structures at coarse-grained level (layouts and formats) is easy to capture, the model can converge quickly, and the introduced fine-grained variabilities (locations and fonts) can bring robustness to the model. A synthesized image may have multiple text blocks, tables and signatures. The size, location, font and rotation of each text area are all random. In this way, we are able to obtain a large-scale labeled dataset with 80,000 samples. We split the dataset with ratio 70\%/15\%/15\% for training/validation/test respectively. Similar to the true dataset, we paste 0\textasciitilde5 scanned signatures into the white space of synthesized image. The specific statistics are listed in table \ref{stat sig}.

\begin{table}
	\caption{Statistics of synthetic document images}
	\label{stat sig}
	\centering
	\begin{tabular}{cccccc}
		\toprule
		\textbf{\# Signatures} & 0 & 1 & 2 & 3 & 4\\
		\midrule
		\textbf{\# Images} & 10,000 & 63,939 & 28,520 & 16,784 & 11,688 \\
		\bottomrule
	\end{tabular}
\end{table}

\subsection{Model Settings}
For region detection tasks, the models use Light Head R-CNN architecture, but replace the original ResNet-101 convolutional feature extractor with ResNet-50. The layout detection model is firstly trained on the synthesized data and fine-tuned on the true dataset. In the contrary, the handwriting detection model is trained only on synthesized images. During inference, the confidence threshold is set to 0.5 for handwriting and 0.1 for other regions. For text recognition tasks, the CNN architecture uses BN-Inception with the final pooling layer and fully connection layers removed. The dimensions of sequential visual features and lstm hidden states are 1024 and 256 respectively. Both hyper-parameter $\alpha$ (Equation \ref{eqn_dist_loss}) and $\lambda$ (Equation \ref{eqn_multi}) are set to 1.
Early stopping and Stochastic Gradient Descent (SGD) with momentum term of 0.9 is employed to train all models. The initial learning rate is 0.001 and the batch size is 32 for all neural network models.

\subsection{Metrics}
We employ multiple metrics for experimental evaluation of different tasks. 

\paragraph{\rm \textbf{Region Detection}}
For region detection models, we use precision and recall score at certain Intersection-over-Union (IoU) threshold as the metrics. IoU represents the overlap ratio between the generated candidate box and the ground truth area, calculated as the intersection area over the union area. A predicted label is regarded as correct if IoU exceeds a certain threshold. In this paper, the IoU threshold is set to 0.85 which is stricter than the common settings. Formally, the precision and recall is calculated by
\begin{equation}
	Precision=\frac{TP}{TP+FP},\quad Recall=\frac{TP}{TP+FN},
\end{equation}
where $TP$, $FP$, $FN$ are \emph{true positive}, \emph{false positive}, and \emph{false negative} predicted labels. 

\paragraph{\rm \textbf{Handwriting Verification}}
For handwriting verification, we use top-1 and top-5 error rates for performance evaluation. Specifically, in the case of top-$n$, the model gives $n$ candidates with highest probabilities and the answer is regarded as correct if the ground truth label is in the $n$ predictions. 

\paragraph{\rm \textbf{Text Recognition}}
For text recognition, we evaluate the performance of recognition with fragment accuracy, which provides the percentage of score fragments correctly recognized. Specifically, accuracy is computed as:
\begin{equation}
Accuracy=\frac{TP + TN}{TP+FP+FN+TN},
\end{equation}
where $TP$, $TN$, $FP$, $FN$ are \emph{true positive}, \emph{true negative}, \emph{false positive}, and \emph{false negative} recognized words.

\begin{table*}
	\caption{Main results of region detection in document images from real scenes.}
	\label{region_detection}
	\centering
	\begin{tabular}{cccccccccc}
		\toprule
		\multirow{2}{1.5cm}{\textbf{Solution}} &
		\multicolumn{2}{c}{\textbf{Handwriting}} &
		\multicolumn{2}{c}{\textbf{Table}} &
		\multicolumn{2}{c}{\textbf{Cell}} &
		\multicolumn{2}{c}{\textbf{Text Block}} &
		\multirow{2}{1.5cm}{\textbf{Inference Time}} \\
		\cline{2-9}
		& Precision & Recall & Precision & Recall &
		Precision & Recall & Precision & Recall & \\
		\midrule
		Adlib PDF & 99.9 & 17.7 & 66.2 & 4.1 & 53.9 & 9.7 & 77.5 & 96.9 & -\\
		Fast CNN  & - & - & 89.6 & 11.2 &  - & - & 92.1 & 91.7 & 0.68 secs\\
		Faster R-CNN & 99.0 & 99.9 & 96.2 & 99.0 & 99.9 & 93.1 & 99.1 & 97.1 & 3.84 secs \\
		\midrule
		Ours &  99.0 & 99.8 & 95.1 & 98.8  & 99.9 & 97.4 & 98.9 & 97.8 & 0.79 secs\\
		\bottomrule
	\end{tabular}
\end{table*}

\subsection{Results and Analyses}
\paragraph{\rm \textbf{Region Detection}}
For the region detection tasks, we compare our models with following SOTA solutions:

\begin{itemize}
	\item \textbf{Adlib OCR}: Adlib OCR\footnote{https://www.adlibsoftware.com} is a enterprise optical character recognition software. It can convert document images into searchable PDF files. Here, tables and text blocks are parsed by pdfplumber\footnote{https://pypi.org/project/pdfplumber/}
	\item \textbf{Fast CNN}: Fast CNN \cite{augusto2017fast} is Deep CNN-based solution for document segmentation and content classification.
	\item \textbf{Faster R-CNN}: Faster R-CNN \cite{ren2015faster} is the SOTA object detection model. \cite{staar2018corpus} proposes to leverage Faster R-CNN for region detection task from document images. We train such a model sharing the same settings with our solutions.
\end{itemize}

The main results are listed in Table \ref{region_detection}. From this table, we can observe that our proposed model outperforms state-for-art solutions on document layout understanding. Adlib PDF and Fast CNN can hardly detect tables and cells since most of the collected images do not contain tabular lines. The performance of Faster R-CNN is competitive with ours. However, it takes almost 5 times longer for inference. 

From Table \ref{warm_up}, the performance drops sharply when training on the \emph{true data} directly while the warm-up strategy brings about 25\% improvements in average. The reason is that the real scene documents has a large amount of complex drawings, making the CNN architecture difficult to map the original image into the latent space and this is the first lesson we learned during model training. Moreover, it should be noted that it is interesting to observe that the test performance of handwriting detection is good enough even only the synthesized images are provided for training. The possible reason is that the features of handwritings significantly differ from the printed texts, which can be captured easily by CNN feature extractors. Another important lesson we learned from this task is that training the model to detect cells and tables jointly can indeed improve the detection accuracy as shown in Table \ref{variant}.

\begin{table}
	\caption{Performance on true data test set with different training sets.}
	\label{warm_up}
	\centering
	\begin{tabular}{ccccc}
		\toprule
		\multirow{2}{0.8cm}{\textbf{Dataset}} &  \multicolumn{4}{c}{\textbf{F-measure}}\\
		\cline{2-5}
		& \textbf{Text} & \textbf{Table} & \textbf{Cell} & \textbf{Handwriting}\\
		\midrule
		Synthetic Data & 58.4 & 49.6 & 51.2 & 99.4 \\
		True Data & 83.1 & 76.3 & 77.0 & 99.2 \\
		Synthetic $\rightarrow$ True & 98.3 & 96.4 & 98.6 & 99.4 \\
		\bottomrule
	\end{tabular}
\end{table}

\begin{table}
	\caption{Performance of variant models on layout detection tasks.}
	\label{variant}
	\centering
	\begin{tabular}{ccccc}
		\toprule
		\multirow{2}{0.1cm}{\#} & \multirow{2}{0.8cm}{\textbf{Model}} &
		\multicolumn{3}{c}{\textbf{F-measure}} \\
		\cline{3-5}
		& & \textbf{Text} & \textbf{Table} & \textbf{Cell}\\
		\midrule
		1 & Light Head R-CNN & 98.3 & 96.4 & 98.6 \\
		2 & Model \#1 without RC (Algorithm \ref{algorithm2}) & 91.8 & 88.3 & 79.4 \\
		3 & Model \#1 with class \emph{table} deleted & 84.6 & - & 17.5 \\
		\bottomrule
	\end{tabular}
\end{table}

\paragraph{\rm \textbf{Text Recognition}}
For text recognition, we compare our models with the SOTA system proposed by \cite{le2019deep}, which combines CTPN and an attention-based encoder decoder model (AED). We refer this system as CTPN-AED for convenience. Comparisons between recognizing accuracy of different models are listed in Table \ref{ocr}. The performance of ACRNN is comparable with CTPN-AED but only needs about half the time for inference. For handwriting recognition, both ACRNN and CTPN-AED achieve quite low scores while the incorporation of handwriting verification in ACRNN-MT brings a significant improvement. This is also an interesting lesson we learned during system designing, that learning to distinguish handwriting features enjoys the ability to boost recognition performance.

\begin{table}
	\caption{Recognition accuracy of different solutions.}
	\label{ocr}
	\begin{tabular}{ccccc}
		\toprule
		\multirow{2}{1cm}{\textbf{Solution}} & \multicolumn{2}{c}{\textbf{Printed Text}} &
		\multicolumn{2}{c}{\textbf{Handwriting}} \\
		\cline{2-5}
		& Accuracy & Speed & Accuracy & Speed \\
		\midrule
		CTPN-AED & 92.2 & 4.2 secs & 71.8 & 2.7 secs \\
		\midrule
		\multicolumn{5}{c}{This work}\\
		\midrule
		ACRNN & 92.7 & 2.5 secs & 71.3 & 1.1 secs \\
		ACRNN-MT & - & - & 88.6 & 1.2 secs\\
		\bottomrule
	\end{tabular}
\end{table}

We also list the results of handwriting verification in Table \ref{handwriting}. We can observe that the Angular Softmax model significant outperforms the conventional softmax classification model and the final Top-5 precision of 97.54 guarantees the practicability of the verification model for industrial application. 

\begin{table}
	\caption{Performance of handwriting verification}
	\label{handwriting}
	\centering
	\begin{tabular}{ccc}
		\toprule
		\textbf{Output Layer} & \textbf{Top-1 error rate} & \textbf{Top-5 error rate}\\
		\midrule
		Softmax & 19.06 & 8.57 \\
		A-Softmax & 9.48 & 2.46 \\
		\bottomrule
	\end{tabular}
\end{table}

\paragraph{\rm \textbf{Incremental Learning}}
It is hard to evaluate the effectiveness of the incremental learning methods since it should be tested in the production environment for a long time, which involves a lot of manual effort. We leave it as an open question for future study. In this paper, we conduct a simulated experiment and, in one sense, prove the effectiveness of the method. To be specific, we artificially synthesize a dataset of 500 images containing short text sequences, and split the data as 400/50/50 for training/validation/text respectively. These images then appended to the signature dataset as a new class. For convenience, we denote the set the original 486 classes as partition A and the new coming class as B. The Top-1 error rate of handwriting classification is listed in Table \ref{incre}. The baseline in Row 1 is fine-tuned directly only on the new data. According to the results, the model fine-tuned on the new class suffers from performance degradation on the original classes. In contrary, the classification precision of the model trained with Algorithm \ref{algorithm1} is similar to the old model in Table \ref{handwriting}, indicating the effectiveness of our method.

\begin{table}
	\caption{Top-1 precison of handwriting verification when a new class is added.}
	\label{incre}
	\begin{tabular}{ccc}
		\toprule
		\multirow{2}{1.5cm}{\textbf{Method}} & \multicolumn{2}{c}{\textbf{Top-1 Error Rate}}\\
		\cline{2-3}
		& A(486 classes) & A + B(487 classes)\\
		\midrule
		Fine-tuning on B & 46.80 & 46.56 \\
		Algorithm \ref{algorithm1} & 9.50 & 9.24\\
		\bottomrule
		
	\end{tabular}
\end{table}

\section{Conclusion and Future Work}

In this paper, we present a system that automatically parses document images into the structured formatted data. The system consists of three major functional components: region detection, text recognition and incremental learning module. We conduct experiments on various tasks, and the results indicate the effectiveness and efficiency of our solutions.

An interesting topic for future study is to systematically evaluate the performance of our system in the production environment. We are also interested in investigating approaches to extract data from natural language sentences in document files.

\bibliographystyle{ACM-Reference-Format}
\bibliography{kdd}

\end{document}